# MC-MLP：Multiple Coordinate Frames in all-MLP Architecture for Vision

Zhimin Zhu[1], Jianguo Zhao[1], Tong Mu[1], Yuliang Yang[1*], Mengyu Zhu[2]

[1]School of Computer and Communication Engineering, University of Science and Technology Beijing, Beijing, China

[2] School of Life Science, Beijing Institute of Technology, Beijing, China

*Corresponding Author: Yuliang Yang, Email: yangbit@ustb.edu.cn

## Abstract

In deep learning, Multi-Layer Perceptrons (MLPs) have once again garnered attention from researchers. This paper introduces MC-MLP, a general MLP-like backbone for computer vision that is composed of a series of fully-connected (FC) layers. In MC-MLP, we propose that the same semantic information has varying levels of difficulty in learning, depending on the coordinate frame of features. To address this, we perform an orthogonal transform on the feature information, equivalent to changing the coordinate frame of features. Through this design, MC-MLP is equipped with multi-coordinate frame receptive fields and the ability to learn information across different coordinate frames. Experiments demonstrate that MC-MLP outperforms most MLPs in image classification tasks, achieving better performance at the same parameter level. The code will be available at: https://github.com/ZZM11/MC-MLP.

## 1 Introduction

Recently, Transformers have achieved remarkable success in various computer vision tasks[22][24][25][27][29][31][33][35]. The Multi-head self-attention(MHSA) mechanism has powerful information learning ability and can model the long-term dependencies. Therefore, many network models based on MHSA have been proposed. However, most deep learning models focus on acquiring information from the spatial domain[1][2][3][5], or a single coordinate frame. Tang et al.[15][17][18] found that information can be learned through frequency transform and this method significantly improves the performance of the model. Frequency transform can also be considered as a coordinate transform.

We speculate that coordinate frames may affect the learning efficiency. Semantic information is highly abstract in deep neural networks(DNN) and we can not be sure that all information will be learned well under one coordinate frame. It is more reasonable that some semantic information may be more easily learned under one coordinate frame and some under another. One motivation is the coordinate frame. Dr. Hinton also mentioned in an interview of the Wired magazine that coordinate frames play a rather important role in making visual observations perception for humans[38]. His words inspire us to use orthogonal transform to change the coordinate frame.

Orthogonal transforms mean changing the coordinate frames of features. DFT, DCT and Hadamard Transform are common orthogonal transform. In the transform domain, some information can be more effectively learned than in the spatial domain. According to this idea, we abstract the whole model into a general architecture called the multi-coordinate frame network, which does not adopt the MHSA. Our multi-coordinate mechanism is shown in Figure 1.

In this paper, we propose a new MLP block, named MC-Block, based on the multi-coordinate frame. By utilizing MC-Blocks, we construct a concise network called MC-MLP. Unlike complex MHSA structures, MC-MLP only requires a unique multiple coordinate frame structure and full connection layer to achieve the expected performance. We introduce a novel MLP in this work, where we employ Hadamard Transform and DCT to transform features from the spatial domain to the transform domain. By leveraging the concatenation and MLP operations, features can be more efficiently integrated from different domains. The transform domain features are the beneficial supplement of the spatial domain features. Such cross-domain information interaction can improve the learning ability of the model.

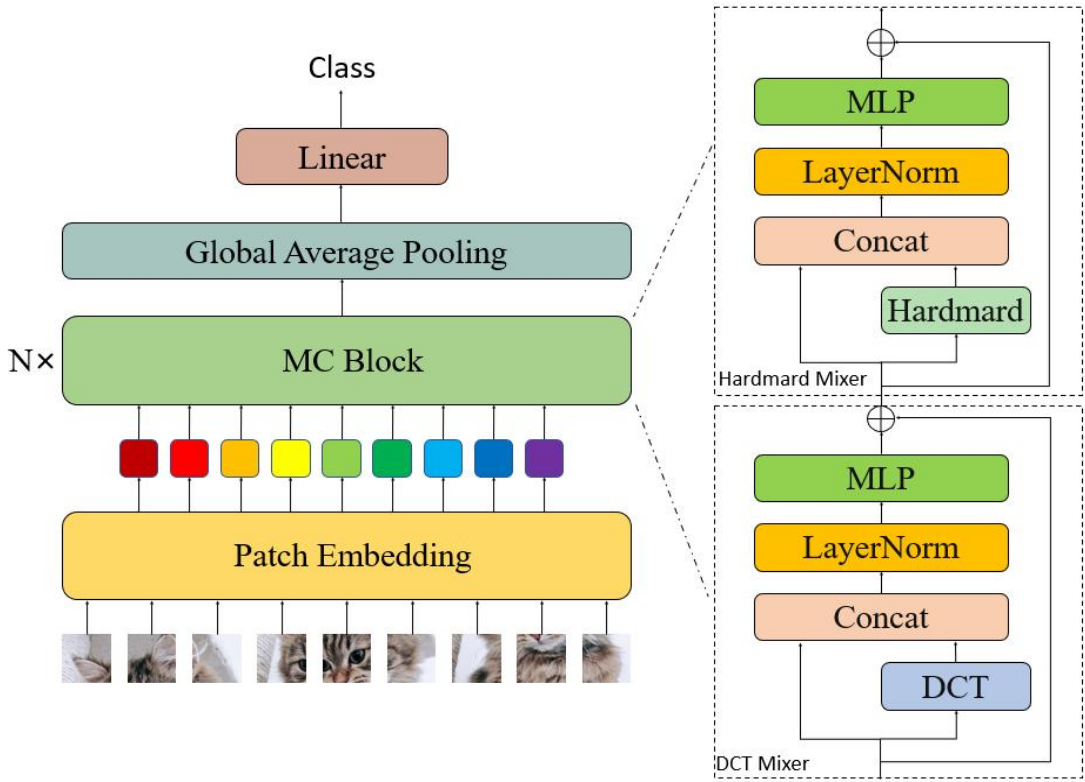

Figure 1: The overall architecture of MC-MLP. Our architecture is based on the MLP and the multi-coordinate frame transform, this layer consists of two key operations: Converts the input spatial features to the transform domain using the 2D DCT and Hadamard Transform, and concatenate the spatial features with the transform domain features. These two steps enable our model to learn in both spatial and transform domain.[26][30]

We use the DCT and Hadamard Transform as the method of multiple coordinate transform, and transform the spatial feature information into the transform domain. The transform domain information will not be limited by the spatial channels and can provide cross-channel information interaction. After obtaining the transform domain features, we concatenate the transform domain features and spatial domain features in the channel dimension, and feed into the full connection layer. With the joint learning and interaction of multi-domain feature information, the potential of our model can be fully unleashed in the moderate parameters and complexity.

In a nutshell, we present a novel MLP-like model design approach that leverages multi-coordinate transform. This method effectively enhances the model's performance by incorporating multi-domain transform, providing more comprehensive understanding of features. By utilizing this approach, our model is able to improve its performance without introducing a complex architecture and the transform adopted in our model has an efficient and fast algorithm.

Our experiments verified the MC-MLP's effectiveness on the CIFAR100 dataset[36]. The results showed that MC-MLP outperformed the recent ViT and MLP models, including ResMLP[3], GFNet[15], and gMLP[2], under similar parameters and FLOPs. These findings indicate that MC-MLP can serve as a viable alternative to MLP models and CNNs in terms of efficiency, generalization ability, and robustness.

Our contributions are summarized as follows:

- Unlike many recent works that use sparse attentions to improve transformer models, we find that orthogonal transforms, along with the standard FC layer, can model various relationships in images.
- We propose a module design method that implements multi-coordinate transform. By incorporating DCT and Hadamard transforms as coordinate transforms, our model learn information from different coordinates to improve the performance. Inverse transforms are not required.
- Based on the proposed MC-Block, we introduce a concise model, called MC-MLP, that consists of a number of basic blocks and attains competitive results.

## 2 Related Work

### 2.1 Self-attention based Models

The Transformer model was initially proposed for the Natural Language Processing (NLP)[51]. With its successful application in image tasks, it has shown remarkable results in image classification[43]. Extensive training on large datasets has demonstrated that the Vision Transformer (ViT) network can achieve performance comparable to the traditional Convolutional Neural Network (CNN) models[5]. Building on these advances, the DeiT model improves the learning strategy of the Transformer to better handle large datasets[11]. To address the computational challenges associated with the self-attention structure of the Transformer, researchers have focused on developing more efficient structures or lighter models[7][10][48].

Some have sought to improve the hybrid architecture by incorporating convolutional layers [44][45][46][47]. Currently, scholars are exploring new token mixing mechanism to more efficiently exchange information, which could obviate the necessity for self-attention[54]. Despite Transformers' achievements, their architecture can be intricate, prompting a growing interest in more streamlined models[7][10][48][55].

## 2.2 MLP-like Models

Recently, MLP models have garnered renewed attention from researchers in the field of computer vision[1][2][3][17][39][40][41][42][50][56]. The MLP-Mixer proposed by Google has demonstrated the potential of MLP models to become the next research paradigm, replacing the traditional CNNs and Transformers. MLP-Mixer and ResMLP both abandon the convolution and attention mechanism, relying solely on stacking two MLP layers: token-mixing MLP and channel-mixing MLP, while still achieving performance comparable to existing mainstream models[1][3]. ViP and sMLP enhance the efficiency and functionality of MLPs by encoding feature representations along two axial dimensions[14][39]. Another research trend involves building a complex receptive field through token mixing to improve the performance of MLP models[40][41][42]. For example, gMLP incorporates Spatial Gated Units (SGUs) to model attention mechanisms, resulting in competitive performance[2]. In Wave MLP, tokens are represented as wave functions with amplitude and phase, and a new token aggregation mechanism is proposed[17]. GFNet is a MLP-like architecture that leverages DFT or FFT, resulting in substantial performance gains[15]. This highlights the potential of orthogonal transform for further exploration. Our proposed MC-MLP offers a novel token aggregation mechanism, which combines token mixing with the coordinate frame transform, providing a more flexible global modeling.

## 3 Model

Figure 1 shows the overall architecture of MC-MLP. In summary, MC-MLP accepts a series of patch embeddings and keeps theirs feature dimension unchanged. It then applies N layers of MC-blocks, each consisting of a Hadamard mixer block and a DCT mixer block. In this section, we introduce our design and describe each component in details.

In this paper, feature maps are represented by tensor $X \in R^{B \times N \times C}$, where $B$ is the batch size, $C$ is the sequence dimension, and $N$ is the sequence length. First, we use a 2D orthogonal transform to transform the input 2D features:

$$Y = F(X) \quad (1)$$

where $F$ represents a two-dimensional(2D) orthogonal transform. In order to correctly complete the transform, it must be ensured that the sequence length ($N$) and sequence dimension ($C$) are integer powers of 2. The 2D orthogonal transform does not change the dimension of the feature vector, where the transformed vector $Y \in R^{B \times N \times C}$. Concatenating the original features and the transformed features, the new dimension is decided by the sequence $Z$:

$$Z = Y \oplus X \tag{2}$$

where $Z \in R^{B \times N \times 2C}$.

Finally, sequence $Z$ feeds into a full connection layer where $LN(Z) = Z'$:

$$Z'' = MLP(Z') + X \tag{3}$$

where LN is Layer Normalization, and MLP is a multi-layer perceptron. MLP can be formulated as follows:

$$MLP(Z') = \sigma(W_2 \sigma(W_1 Z')) \tag{4}$$

where $W_1 \in R^{(f \times n) \times n}, W_2 \in R^{n \times (f \times n)}$. $f \times n$ means the width of the hidden layer.

Based on the aforementioned analysis, we propose two fundamental modules, namely the DCT Mixer and the Hadamard Mixer. These design paradigms are intended to improve the feature extraction and token aggregation capabilities of MLP models, thereby leading to the enhanced performance. By applying either a DCT or Hadamard transform to the input features, the transformed features can be concatenated and further processed via an MLP. The combination of orthogonal transform and MLP can efficiently extract and utilize feature information. The design of our MC-MLP holds significant potential to substantially enhance the performance of MLP models and can serve as a crucial component in future MLP-based models.

## 3.1 DCT Mixer

The DCT Mixer is a critical component of the MC-block. It utilizes the 2D DCT to convert the input features into another coordinate frame, effectively highlighting diverse and complementary information. The transformed features are then concatenated with those in the original domain and processed through a MLP for learning. This approach allows the MLP learning features represented from multiple coordinate frames, leading to improved performance. The implementation of the 2D DCT is achieved through two sequential one-dimensional(1D) DCT.

The 1D DCT can be perceived as the DFT of the input signal treated as a real even function. To convert the input signal to a real even function, it extends any input signal to a real even signal before applying DFT.

The formula for 1D DCT is as follows[57]:

$$F(k) = C(k)\sqrt{\frac{2}{N}} \sum_{n=0}^{N-1} f(n) \cos \frac{\pi(2n+1)k}{2N} \tag{5}$$

$$C(k) = \begin{cases} \frac{1}{\sqrt{2}}, & k = 0 \\ 1, & 1 \leq k \leq N-1 \end{cases} \tag{6}$$

Where $f(n)$ is a discrete signal sequence. $C(k)$ is a constant determined by $k$.

For 2D discrete sequences such as images, let $F(i,j)$ be the image matrix of $M \times N$, and its 2D DCT formula is as follows:

$$F(u,v) = c(u)c(v) \sum_{i=0}^{N-1} \cdot \sum_{j=0}^{N-1} f(i,j) \cos\left[\frac{(i+0.5)\pi}{N}u\right] \cos\left[\frac{(j+0.5)\pi}{N}v\right] \tag{7}$$

$$c(u) = \begin{cases} \sqrt{\frac{1}{N}}, & u = 0 \\ \sqrt{\frac{2}{N}}, & u \neq 0 \end{cases} \quad (8)$$

Where $0 \leq i \leq N, 0 \leq j \leq M$.

The specific calculation amount varies slightly depending on the specific algorithms, and the magnitude of the computation is $O(MNlog(MN))$[52]. The specific process is illustrated in the figure below:

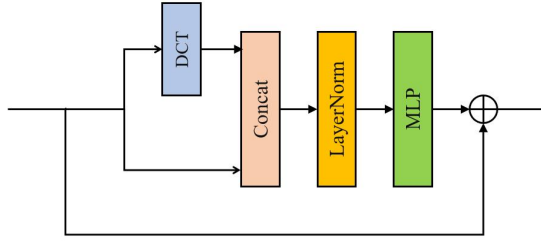

Figure 2: The structure of DCT-Mixer.

## 3.2 Hadamard Mixer

The Hadamard[53] mixer can be used as a complementary module to the DCT mixer, and it further enhances the overall performance of the model. This module embodies the multi-coordinate frame learning principle introduced in this study. In principle, the 2D Hadamard transform is achieved by applying two separate 2D Hadamard matrices to the original feature matrix, one as a left-multiplication and the other as a right-multiplication. The fast Hadamard algorithm can significantly reduce the amount of computation.

The Hadamard transform is a non-sinusoidal transform that utilizes the Hadamard matrix as its transform formula. It concentrates information on the characteristic corners after transform. The Hadamard transform only requires addition or subtraction, without multiplication, resulting in low computational cost. The Hadamard matrix is an orthogonal matrix that consists of +1 and -1 elements. Its first-order Hadamard matrix is defined as:

$$H_1 = [1] \quad (9)$$

The second order Hadamard matrix is defined as

$$H_2 = \begin{bmatrix} 1 & 1 \\ 1 & -1 \end{bmatrix} \quad (10)$$

Higher order Hadamard matrix can be derived from lower order, namely:

$$H_N = H_{2^n} = H_2 \otimes H_{2^{n-1}} = \begin{bmatrix} H_{2^{n-1}} & H_{2^{n-1}} \\ H_{2^{n-1}} & H_{2^{n-1}} \end{bmatrix} = \begin{bmatrix} H_{\frac{N}{2}} & H_{\frac{N}{2}} \\ H_{\frac{N}{2}} & -H_{\frac{N}{2}} \end{bmatrix} \quad (11)$$

The 2D Hadamard transform involves left-multiplying the original matrix by a Hadamard matrix of corresponding order and right-multiplying it by another Hadamard matrix of the same order, followed by division by the square of the order. The process is illustrated below:

$$W = \frac{H_n \otimes X \otimes H_n}{n^2} \quad (12)$$

Assuming that the dimensions of the input 2D data are $M \times N$, where $M$ and $N$ are both integer powers of 2, the computation of the Hadamard Fast Algorithm is $O(MNlog(MN))$ .The specific process is shown in the figure:

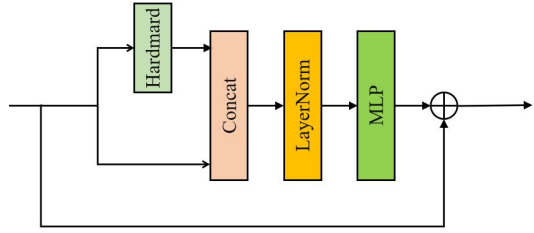

Figure 3: The structure of Hardmard Mixer.

# 4  Experiments

We utilized the PyTorch and TIMM code base to implement our model and trained it on two NVIDIA Titan XP graphics cards. The training was conducted for 300 epochs from scratch on the CIFAR-100[36] dataset. We employed the AdamW optimizer[37] and set the initial learning rate to 0.01. The learning rate attenuation strategy employed was cosine attenuation with linear preheating. Throughout the training and testing phases, the input model image size remained fixed at 224 × 224.

The CIFAR-100[36] dataset comprises 60,000 32 × 32 color images containing 600 images each. Each category includes 500 training images and 100 test images. During model training, we employed the Mixup and Cutmix data enhancement techniques. The specific super parameters for training are presented in the following figure:

Table 1：Hyperparameters for Image classification on CIFAR-100.

| | |
|---|---|
| lr scheduler | cosine |
| Epochs | 300 |
| Warmup epoch | 3 |
| lr | 0.01 |
| Optimizer | AdamW |
| Weight decay | 1×e$^{-5}$ |
| Cutmix | 0.4 |
| Mixup | 0.2 |

The accuracy of Top-1 trained from scratch on CIFAR-100[36] is 78.64%. The model parameters and calculation amount are as follows:

Table 2: Experiments on CIFAR-100 without pre-training.

| Arch | Model | Params | GMAC | Top-1(%) |
|---|---|---|---|---|
| CNN | ResNet-50[13] | 25.6 | 4.1 | 75.1 |
| Transfomer | ViT[5] | 22.1 | 4.6 | 66.45 |
| CNN-ViT | Conformer[49] | 23.5 | 5.2 | 75.43 |
| MLP | MLP-Mixer[1] | 28.0 | 4.4 | 62.16 |
| MLP | ViP[14] | 25.0 | 8.5 | 70.51 |
| MLP | ResMLP[3] | 30.0 | 6.0 | 66.40 |
| MLP | AS-MLP[50] | 28.0 | 4.4 | 65.16 |
| MLP | gMLP[2] | 24.5 | 5.56 | 64.80 |
| MLP | MC-MLP(ours) | 25.8 | 6.0 | 78.64 |

# 5 Conclusion

This paper introduces the MC-MLP architecture, which is a computationally efficient image classification model for vision tasks. Our proposed MC-block replaces the self-attention sublayer with a combination of DCT and Hadamard transform. Transform completes the transformation of the coordinate frame of features, where the difficulty in learning features can vary across domains. Moreover, the computational cost of DCT and Hadamard transform is low, as fast algorithms exist.

Our architecture is highly efficient, benefiting from the multi-coordinate transform. Our experimental results demonstrate that MC-MLP is a competitive alternative to visual transformers, MLP-like models, and CNNs in terms of the accuracy/complexity trade-off. In the future, we will investigate the potential of MLP-like structures in different tasks further.